\documentclass[times,3p,preprint]{elsarticle}

\usepackage{lineno,hyperref}
\modulolinenumbers[5]

\journal{Journal of Computational Physics}

\usepackage{amsmath}
\DeclareMathOperator*{\argmin}{argmin}
\newcommand{\sym}[2]{(\protect\includegraphics[height=6pt]{styles/#1_#2})}

\begin{document}

\begin{frontmatter}

\title{Nonlinear integro--differential operator regression with neural networks}

\author[Ad3]{Ravi G.\ Patel\corref{cor1}}
\cortext[cor1]{Corresponding author}
\ead{rgp62@cornell.edu}
\address[Ad3]{Sibley School of Mechanical and Aerospace Engineering, Cornell University, Cornell, NY 14853-7501, USA}

\author[Ad3]{Olivier Desjardins}
\ead{olivier.desjardins@cornell.edu}

\begin{abstract}
This note introduces a regression technique for finding a class of nonlinear integro-differential operators from data. The method parametrizes the spatial operator with neural networks and Fourier transforms such that it can fit a class of nonlinear operators without needing a library of \textit{a priori} selected operators. We verify that this method can recover the spatial operators in the fractional heat equation and the Kuramoto--Sivashinsky equation from numerical solutions of the equations.
\end{abstract}


\end{frontmatter}

\section{Introduction}

Machine learning is a promising tool for physical modeling. One potential application is obtaining equations that describe a dataset \cite{Brunton2016,Raissi2018}. Gulian et al. have recently introduced a method for obtaining space--fractional differential equations, finding differential operators in Fourier space \cite{Gulian2018}. However, their method is limited to linear operators. In this note, we introduce a method for finding nonlinear integro--differential equations from data using a combination neural networks and Fourier pseudospectral methods.

\section{Method} \label{sec:method}

Suppose we have sets of temporally evolving spatial data, $u_\alpha\left(x,t\right)$, starting at different initial conditions indexed by $\alpha$. Let us assume this dataset is governed by a first order in time equation,
\begin{equation} \label{eq:pde}
    \partial_t u = \mathcal{N}\left\{u\right\},
\end{equation}
where $\mathcal{N}$ is a nonlinear integro--differential operator. Restricting ourselves to 1--D, periodic domains, a large class of $\mathcal{N}$'s might take the form
\begin{equation} \label{eq:op}
    \mathcal{N}\left\{u\right\} = \sum_\gamma \mathcal{N}_\gamma \left\{u\right\} = \sum_\gamma \mathcal{F}^{-1}\left\{ g_\gamma \left(\kappa\right) \mathcal{F}\left\{ h_\gamma \left(u\right)\right\}\right\},
\end{equation}
where $\mathcal{F}$ is the Fourier transform, $h_\gamma$ are functions of $u$, and $g_\gamma$ are Hermitian functions of the wavenumber, $\kappa$. In practice, we rely on Fast Fourier Transforms (FFTs) and Inverse FFTs specialized for real data, implicitly assuming Hermitian symmetry for $g_\gamma$. As an example, consider the viscous Burgers' equation as a special case of Eq.~\ref{eq:op},
\begin{equation} \label{eq:burg}
    \partial_t u = - \frac{1}{2}\partial_x u^2 + \partial_x^2 u = \mathcal{F}^{-1}\left\{-\frac{i\kappa}{2} \mathcal{F}\left\{ u^2\right\}\right\} + \mathcal{F}^{-1}\left\{ - \kappa^2 \mathcal{F}\left\{ u\right\}\right\}.
\end{equation}
Many other physically relevant PDEs also take the form of Eq.~\ref{eq:op} in periodic domains. In addition, the spatial operators in Eq.~\ref{eq:op} may efficiently be evaluated numerically via pseudospectral methods. Therefore, seeking operators of this form may be useful and practical for physical modeling.

The explicit, first order update of $u$ given by Eq.~\ref{eq:pde} is $u^{n+1} = \left( I + \Delta t \mathcal{N}\right) \left\{u^n\right\}$, where $I$ is the identity operator, $n$ is the time step index and $\Delta t$ is the time step size. Given example pairs of $u_\alpha^{n+1}$ and $u_\alpha^{n}$, we seek functions $g_\gamma$ and $h_\gamma$ that minimize the error
\begin{equation} \label{eq:reg}
    \argmin_{g_\gamma,h_\gamma} \sum_{n,\alpha} { \int{ \left( u_\alpha^{n+1} - \left(u_\alpha^n + \Delta t\sum_\gamma \mathcal{F}^{-1}\left\{ g_\gamma \left(\kappa\right) \mathcal{F}\left\{ h_\gamma \left(u_\alpha^n\right)\right\}\right\} \right)\right)^2 dx}},
\end{equation}
thereby specifying a $\mathcal{N}$. We use $m$--layer neural networks to represent $g_\gamma$ and $h_\gamma$,
\begin{align} \label{eq:nn}
    &g_\gamma\left(\kappa\right) =  \sum_z w_{g_{\gamma;z}}^m  \hdots  \sum_i \sigma\left(w_{g_{\gamma;ji}}^1 \sigma \left( w_{{g_{\gamma;i}}}^0 \kappa + b_{g_{\gamma;i}}^0 \right)+b_{g_{\gamma;j}}^1\right) \hdots + b_{g_\gamma}^m \\
    &h_\gamma\left(u\right) =  \sum_z w_{h_{\gamma;z}}^m \hdots  \sum_i \sigma\left(w_{h_{\gamma;ji}}^1 \sigma \left( w_{{h_{\gamma;i}}}^0 u + b_{h_{\gamma;i}}^0 \right)+b_{h_{\gamma;j}}^1\right) \hdots + b_{h_\gamma}^m
\end{align}
where $\sigma$ is an activation function, applied element--wise, and $w$ and $b$ are the weights and biases to be found via Adam optimization \cite{Kingma2014}. For many practical problems, neural network regression has been shown to be efficient and capable of converging to good minima. 

We can employ a few additional techniques to convergence. We may require more stringently that $\mathcal{N}$ be consistent with multiple updates. For $p$ updates, $u^{n+p} =\left( I + \Delta t \mathcal{N}\right)^p \left\{u^n\right\}$, the equivalent regression problem to Eq.~\ref{eq:reg} can be specified. For many physical systems, we may expect a conservation equation, i.e.,
\begin{equation}\label{eq:regcons}
    \partial_t u = \mathcal{N} \left\{ u \right\} = \partial_x \hat{\mathcal{N}} \left\{ u \right\} =  \sum_\gamma \mathcal{F}^{-1}\left\{ i\kappa \hat{g}_\gamma \left(\kappa\right) \mathcal{F}\left\{ h_\gamma \left(u\right)\right\}\right\}.
\end{equation}
where we have absorbed the spatial derivative into Fourier space so that $g_\gamma\left(\kappa\right)=i\kappa \hat{g}_\gamma\left(\kappa\right)$. We may also expect some symmetry under coordinate reflection. We can force $h\left( u\right)$ to be either odd, $h\left(u\right) = \mathrm{sign}\left(u\right) \hat{h}\left(\left|u\right|\right)$, or even, $h\left(u\right) =  \hat{h}\left(\left|u\right|\right)$. Finally, we may restrict $g_\gamma\left( \kappa\right)$ to be either purely real or purely imaginary.

\section{Results}

We test the ability of this method to recover approximations of two known equations from simulations of those equations. For each equation we perform the following:

\begin{enumerate}
    \item We generate a few different initial conditions using low pass filtered white noise. Using these initial conditions, we numerically integrate the original equation with first order, explicit updates, evaluating all spatial operators pseudospectrally.
    \item We use the method outlined in Sec.~\ref{sec:method} on the simulated data to find an equation in the form of Eq.~\ref{eq:pde}. We use two hidden layers with five nodes each for the $g_\gamma$ and $h_\gamma$ in each $\mathcal{N}_\gamma$. We use the exponential linear unit for the activation functions \cite{Clevert2015}. We exclusively seek operators of the form given by Eq.~\ref{eq:regcons}. We seek four operators, one with real $g_\gamma$ and even $h_\gamma$, one with real $g_\gamma$ and odd $h_\gamma$, one with imaginary $g_\gamma$ and even $h_\gamma$, and one with imaginary $g_\gamma$ and odd $h_\gamma$. We perform the regression ten times, increasing the number of updates, $p$, by one for each successive iteration.
    \item We generate a new initial condition. For comparison purposes, we numerically integrate the original equation and the equation found from regression using this new initial condition.
\end{enumerate}
To prevent aliasing \cite{Canuto1988}, we use $n=192$ collocation points, but only use wavenumbers $\left|\kappa \leq n/3 \right|$ and zero out $\mathcal{F}\left\{ h_\gamma \left( u \right) \right\}$ for larger wavenumbers. We implement both the numerical method and the regression tool in TensorFlow \cite{tensorflow2015-whitepaper}.

\subsection{Fractional heat equation}

We first test this regression technique against the fractional heat equation,
\begin{equation}\label{eq:heat}
    \partial_t u = 0.01\nabla^{3/4} u,
\end{equation}
where $\nabla^{3/4}$ is a fractional Laplacian of order $\frac{3}{4}$ and defined via Fourier transform as \cite{Raissi2018,Gulian2018}
\begin{equation}
    \mathcal{F} \left\{ \nabla^{3/4} u \right\} = -\left| \kappa \right|^{3/2}\mathcal{F} \left\{ u \right\}.
\end{equation}

\begin{figure}[h]
\includegraphics[width=6.5in]{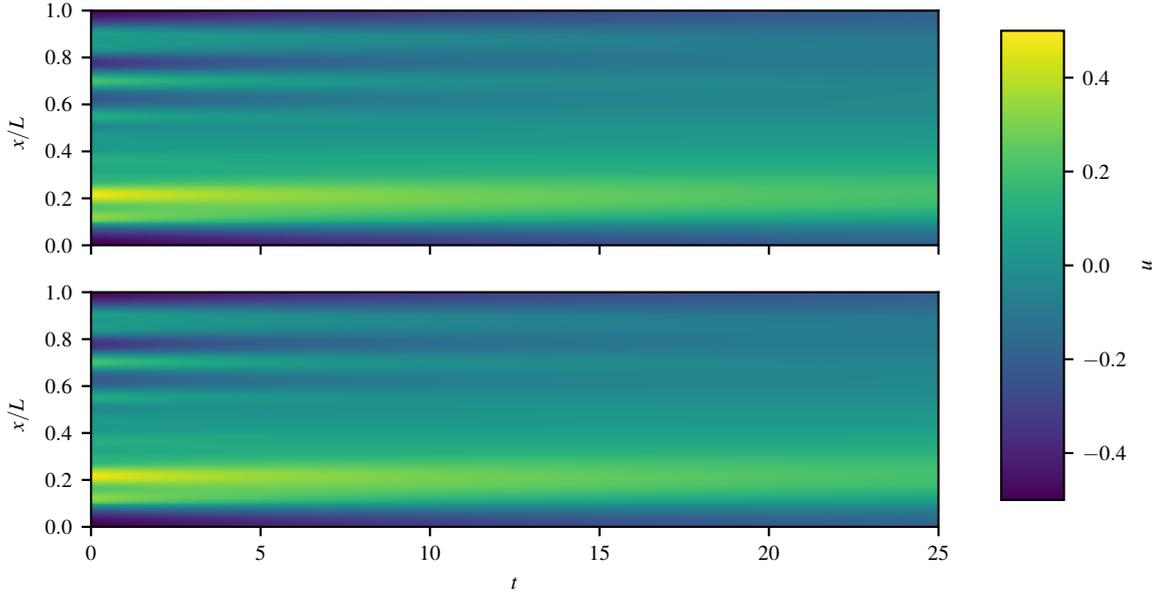}
\caption{Evolution of the fractional heat equation with $L=2\pi$. Original equation (top) and regressed equation (bottom).}
\label{fig:heat}
\end{figure}

After training, a new initial condition is generated and evolved by both the original equation and the regressed equation. Figure~\ref{fig:heat} demonstrates that the regressed equation produces similar evolution as produced by the original equation.

%
%
%

\subsection{Kuramoto--Sivashinsky equation}\label{sec:ks}

\begin{figure}[h] 
\includegraphics[width=6.5in]{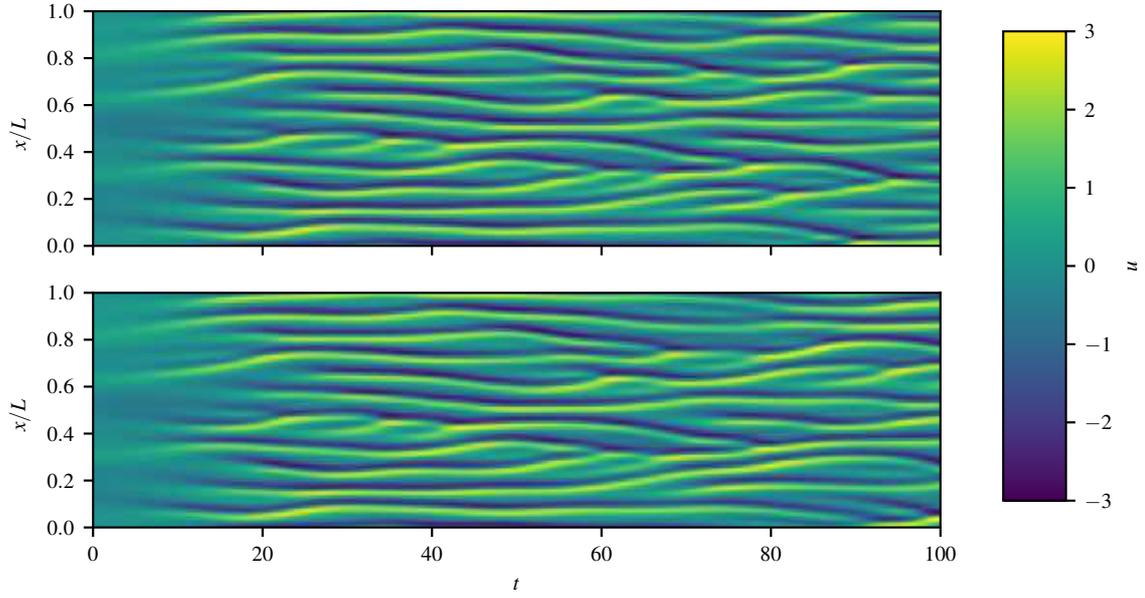}
\caption{Evolution of the Kuramoto--Sivashinsky equation with $L=32\pi$. Original equation (top) and regressed equation (bottom).}
\label{fig:ks}
\end{figure}

The Kuramoto--Sivashinsky is given as \cite{Raissi2018,Rudy2017}
\begin{equation} \label{eq:ks}
    \partial_t u + \frac{1}{2} \partial_x u^2 + \partial_x^{2} u + \partial_x^{4} u=0.
\end{equation}
This equation produces self--sustaining, chaotic dynamics and has previously been used to verify regression techniques for model discovery \cite{Raissi2018,Rudy2017}. Figure~\ref{fig:ks} shows the evolution of the original equation and regressed equation. At early times, the regressed equation matches the original equation, but at later times, the solutions diverge. This can be attributed to the chaotic evolution of the Kuramoto--Sivashinsky equation since small errors in the regressed model result in large differences over time. Qualitatively, the original equation and regressed equation produce similar features even at later times.

\begin{figure}[h] 
\includegraphics[width=6.5in]{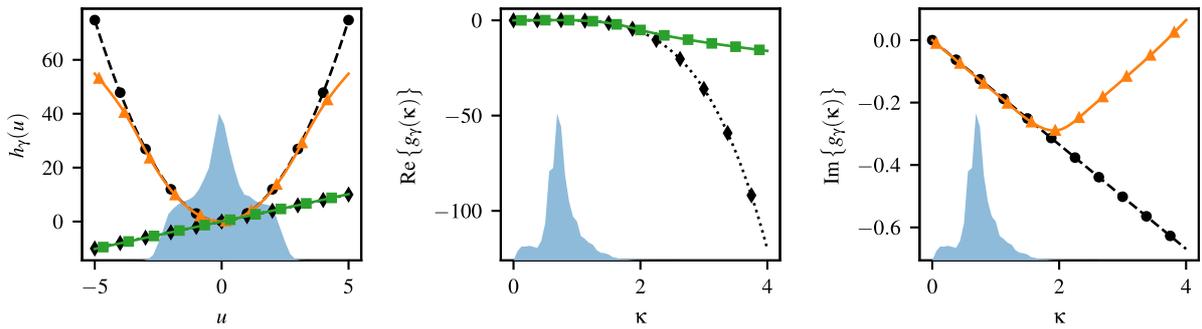}
\caption{Parameterizations of the operators. The original convective operator \sym{o}{k} and original combined inverse diffusion and hyperdiffusion diffusion operator \sym{d}{k} are compared against the regressed operators \sym{t}{C1} and \sym{sq}{C2}. The density of the $u$ samples (left) and the energy spectrum of the $u$ samples (middle and right) are shown by the shaded regions \sym{sh}{C0}. These plots have been normalized for visual clarity.}
\label{fig:op}
\end{figure}

The parameterization of the spatial operator in Eq.~\ref{eq:op} allows for comparison between the original equation and the regressed equation. The Kuramoto--Sivashinsky equation can be written in the form of Eq.~\ref{eq:op} as
\begin{equation} \label{eq:ks_param}
    \partial_t u = \mathcal{F}^{-1} \left\{ \left(\kappa^2 - \kappa^4\right) \mathcal{F} \left\{ u \right\}  \right\} + \mathcal{F}^{-1} \left\{ -\frac{1}{2} i\kappa \mathcal{F} \left\{ u^2 \right\}  \right\}.
\end{equation}
We can compare the parameterizations of the regressed operators to the corresponding functions in Eq.~\ref{eq:ks_param},  $g_1^o= \kappa^2-\kappa^4$, $h_1^o=u$, $g_2^o=-\frac{1}{2}i\kappa$, and $h_2^o=u^2$. Because $h_\gamma$ and $g_\gamma$ are only unique up to a multiplicative constant, $h_\gamma$ is normalized by $\int_0^{1} h_\gamma du$ and the normalization factor multiplied into $g_\gamma$. Figure~\ref{fig:op} compares the parameterizations. The operator with even $h$ and real $g$ and the operator with odd $h$ and imaginary $g$ are dropped from these plots because they are approximately zero. The regression technique produces an identifiable convective operator \sym{t}{C1} and combined inverse diffusion and hyperdiffusion operator \sym{sq}{C2}. For small $|u|$ and small $\kappa$, the regressed $g_\gamma$'s and $h_\gamma$'s  match well with their corresponding $g_\gamma^o$'s and $h_\gamma^o$'s. At large $|u|$ and large $\kappa$, these parameterizations diverge from the true values. The regression technique cannot adequately fit the $g_\gamma$'s and $h_\gamma$'s at these larger values of $|u|$ and $\kappa$ because they are poorly sampled, as suggested by the density and energy spectrum of the samples. However, it is able to produce equations suitable for well--sampled dynamics.

\section{Conclusion}

This note introduces a regression technique for obtaining integro--differential equations from data and verifies it against numerical simulations of the fractional heat equation and the Kuramoto--Sivashinsky equation. We find the regressed equations produce dynamics similar to the original equations. Future work may focus on using the method developed in this note towards closure models of the filtered Burgers' equation, extending the method to systems of equations, or extending the method to multi--dimensional equations.
 
\section*{Acknowledgments}
RGP and OD were supported by a grant from the National Science Foundation (CBET-1437903).

\section*{References}

\bibliography{opreg}

\end{document}